\documentclass{article}

\usepackage{arxiv}

\usepackage[utf8]{inputenc} 
\usepackage[T1]{fontenc}    
\usepackage{hyperref}       
\usepackage{url}            
\usepackage{booktabs}       
\usepackage{amsfonts}       
\usepackage{nicefrac}       
\usepackage{microtype}      
\usepackage{lipsum}

\usepackage{times}
\usepackage{epsfig}
\usepackage{graphicx}
\usepackage{amsmath}
\usepackage{amssymb}

\usepackage{color}
\usepackage{relsize}
\usepackage{tabularx}
\usepackage{wrapfig}
\usepackage{multirow}
\usepackage{multicol}
\usepackage{caption}
\usepackage{subcaption}
\usepackage{booktabs}
\usepackage{float}

\newcommand{\etal}{\textit{et al}.}
\newcommand{\ie}{\textit{i}.\textit{e}. }
\newcommand{\eg}{\textit{e}.\textit{g}. }
\newcommand{\defas}{\overset{\underset{\mathrm{def}}{}}{=}}
\newcommand{\R}{\ensuremath{\mathbb{R}} }
\newcommand{\M}{\ensuremath{\mathcal{M}} }
\newcommand{\g}{\ensuremath{\mathfrak{g}} }

\DeclareMathOperator{\rot}{Rot}

\title{ZerNet: Convolutional Neural Networks on Arbitrary Surfaces \\via Zernike Local Tangent Space Estimation}

\author{
  Zhiyu Sun \\
  Department of Industrial and Systems Engineering\\
  The University of Iowa\\
  \texttt{zhiyu-sun@uiowa.edu} \\
   \And
 Ethan Rooke \\
  Applied Mathematical and Computational Sciences\\
  The University of Iowa\\
  \texttt{ethan-rooke@uiowa.edu} \\
   \And
    Jerome Charton \\
  Department of Industrial and Systems Engineering \\
  The University of Iowa\\
  \texttt{jerome.charton@lab327.net} \\
   \And
 Yusen He \\
  Department of Industrial and Systems Engineering\\
  The University of Iowa\\
  \texttt{yusen-he@uiowa.edu} \\
   \And
 Jia Lu \\
  Department of Mechanical Engineering\\
  The University of Iowa\\
  \texttt{jia-lu@uiowa.edu} \\
   \And
 Stephen Baek\thanks{Corresponding Author: stephen-baek@uiowa.edu} \\
  Department of Industrial and Systems Engineering\\
  The University of Iowa\\
  \texttt{stephen-baek@uiowa.edu} \\
}
\begin{document}
\maketitle


\begin{abstract}
In this paper, we propose a novel formulation to extend CNNs to two-dimensional (2D) manifolds using orthogonal basis functions, called Zernike polynomials.
In many areas, geometric features play a key role in understanding scientific phenomena. Thus, an ability to codify geometric features into a mathematical quantity can be critical. Recently, convolutional neural networks (CNNs) have demonstrated the promising capability of extracting and codifying features from visual information. However, the progress has been concentrated in computer vision applications where there exists an inherent grid-like structure. In contrast, many geometry processing problems are defined on curved surfaces, and the generalization of CNNs is not quite trivial. The difficulties are rooted in the lack of key ingredients such as the canonical grid-like representation, the notion of consistent orientation, and a compatible local topology across the domain. In this paper, we prove that the convolution of two functions can be represented as a simple dot product between Zernike polynomial coefficients; and the rotation of a convolution kernel is essentially a set of $2\times2$ rotation matrices applied to the coefficients. As such, the key contribution of this work resides in a concise but rigorous mathematical generalization of the CNN building blocks.
\end{abstract}


\section{Introduction}
\label{intro}
Many areas of science and engineering have to deal with geometric data because shape often encompasses critical information for understanding the phenomena. Such problems often boil down to identifying latent geometric patterns behind a diversity of shapes and correlating these geometric characteristics with certain physical phenomena.

In this regard, convolutional neural networks (CNNs) have demonstrated a remarkable capacity in capturing and recognizing important visual features from images or signals, conceiving the unprecedented advances over the last half-decade. It is partly due to having canonical grid-like structures granted on regular signal or image domains that the basic operations of CNNs (e.g. convolutions and poolings) could be defined.

The geometric processing community defines many visual recognition problems on discretely-sampled arbitrary surfaces. For instance, segmentation of a LiDAR scan \cite{moosmann2009segmentation,douillard2011segmentation} can be understood as a point-wise classification problem defined on a point-cloud-approximated local surface. Feature detection \cite{song2014sliding,Harik2017:ShapeTerra,nurunnabi2015outlier} and establishing correspondence \cite{Zaharescu2009:meshHOG,Sun2017a,rodola2014dense} on 3D meshes are other typical examples of manifold-based visual recognition problems.

\begin{figure}[h]
    \centering
    \begin{subfigure}[b]{0.3\linewidth}
        \centering
        \includegraphics[width=0.65\linewidth]{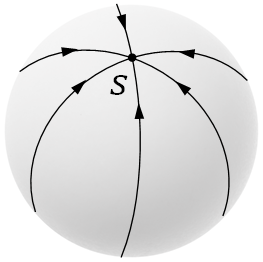}
        \caption{}
        \label{fig:hairyball}
    \end{subfigure}
    \begin{subfigure}[b]{0.3\linewidth}
        \centering
        \includegraphics[width=0.65\linewidth]{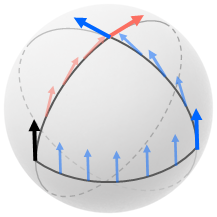}
        \caption{}
        \label{fig:holonomy}
    \end{subfigure}
    \begin{subfigure}[b]{0.37\linewidth}
        \centering
        \includegraphics[width=0.65\linewidth]{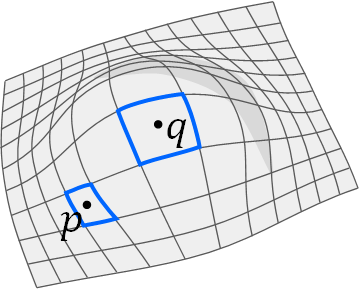}
        \caption{}
        \label{fig:metric_distortion}
    \end{subfigure}
    \caption{Problems associated with manifolds when expanding the notion of convolution. (a) a topological sphere cannot have a grid defined on it without a singularity ($S$); (b) the end-up direction of a vector parallel-transported along a loop on a manifold is path-dependency; (c) irregularity of a grid causes varying-size receptive fields.}
    \label{fig:problems_with_manifold}
\end{figure}

By their nature, most visual recognition problems are analogous to those in the Euclidean setting. In practice, however, extending CNNs to manifolds is not trivial due to the lack of fundamental operations which exist in Euclidean space. As Bronstein \etal~\cite{bronstein2017geometric} point out, ``(CNNs) have been most successful on data with an underlying Euclidean or grid-like structure, and in cases where the invariants of these structures are built into networks used to model them.'' For instance, the hairy ball theorem \cite{eisenberg1979proof} shows that it is impossible to install a grid on a domain homeomorphic to  a sphere without creating a singular point where we lose the notion of orientation (Figure \ref{fig:hairyball}). And even though there was a grid structure somehow defined without a singularity, when we transport a vector in parallel across the manifold, the direction of the vector can end up differently according to the path we take, which is so-called holonomy (Figure \ref{fig:holonomy}). Therefore, the sliding convolution kernels along a loop path on a surface can cause a change of orientation when it comes back to the starting point. Moreover, the grid may be irregular, causing varying sizes of the effective receptive fields across the manifold (Figure \ref{fig:metric_distortion}).

In this research, we propose a new mathematical formulation of CNNs on arbitrary surfaces. The key innovation of the proposed approach resides in the local piece-wise parameterization of a tensor field via the Zernike polynomial bases \cite{von1934beugungstheorie}. These orthogonal basis functions allow a rigorous but straightforward generalization of convolution via a simple dot product between Zernike polynomial coefficients. Furthermore, we simplify kernel rotations to a set of 2D rotation matrices applied to the Zernike polynomial coefficients, enabling concise definition of angular pooling, which is critical on manifold domains. The main contribution of this article is to generalize the representation of CNN operators on surfaces while maintaining their simplicity.
Finally, building upon these new theoretical grounds, we propose a new approach called ZerNet (Zernike CNN), and demonstrate our approach outperforming other state-of-the-art methods in both classification and regression tasks.
\section{Related Works}
\label{relatedworks}

Although there has been vigorous research activity surrounding CNNs, only a few have focused on non-Euclidean CNNs, despite the numerous applications and benefits mentioned above. Existing work on non-Euclidean CNNs falls into one of the two following categories: (1) spectral methods and (2) spatial methods.

\noindent\textbf{Spectral methods} \indent Pioneering works \cite{bruna2013spectral,henaff2015deep,defferrard2016convolutional,yi2017syncspeccnn} defined the convolution operation on manifolds by employing a \textit{spectral graph processing} approach. The main theoretical foundation for this is the convolution theorem, which states that the convolution $f*g$ of two functions $f$ and $g$ is equivalent to a simple element-wise product in the Fourier (spectral) domain: $\mathcal{F}(f*g) = \mathcal{F}(f) \odot \mathcal{F}(g)$, where $\mathcal{F}$ denotes the Fourier transform and $\odot$ is the element-wise Hadamard product. The convolution theorem generalizes to manifolds quite effortlessly as when we let $U$ be the linear Fourier operator on a manifold, the convolution theorem gives:
\begin{equation}
    f * g = U^\top\left\{ (U f) \odot (U g) \right\}.
    \label{eq:spectral_conv}
\end{equation}

The manifold Fourier operator $U$ is essentially the eigenfuctions of the Laplace-Beltrami operator $\Delta_\g$ defined on $(\M, \g)$. The Laplace-Beltrami operator $\Delta_\g$ is a generalization of the second-order derivative, Laplacian, on Riemannian manifolds and the discretization of the Laplace-Beltrami operator is well understood in literature, e.g., as in \cite{meyer2003discrete}.

This elegant generalization suffers from three main flaws in practice. First, the computation of $U$ is prohibitively expensive when there is a large set of vertices, as it requires the eigendecomposition of the linear operator $\Delta_\g$.  More recent works \cite{defferrard2016convolutional,kipf2016semi,boscaini2015learning} have introduced circumventing measures to limit the computation of eigenfunctions to local regions by using computational techniques such as \textit{windowed} Fourier transformations \cite{shuman2016vertex}. 

Aside from these speed concerns, a more fundamental issue lies in the numerical behavior of the eigenfunctions. It is well-known to the geometry processing community that the eigenfunctions can flip signs and change orderings \cite{Levy:2010:SMP,rong2008spectral}. In the context of geometric data analysis, this can be critical since the eigenbases across different geometric models may vary and may not be compatible, making it impossible to represent convolution kernels consistently. Finally, spectral kernels are rotationally-symmetric as the spectral bases are isotropic \cite{bruna2013spectral,boscaini2015learning}, which limits the expressiveness substantially \cite{cohen2019gauge}.

\noindent\textbf{Spatial methods} \indent Spatial approaches follow a more straightforward and explicit generalization of CNNs, as in \cite{duvenaud2015convolutional,niepert2016learning,hechtlinger2017generalization}. In contrast to the spectral formulations where the convolution is implicitly reparameterized using spectral bases, in spatial formulations, convolution is defined more explicitly and intuitively on tangent spaces of the manifold. For example, Masci \etal \cite{masci2015geodesic} applied convolution kernels on local geodesic disks to approximate the tangent spaces. This idea quite naturally generalizes the explicit notion of convolution onto manifold domains. In a similar work, Boscaini \etal \cite{boscaini2016learning} used anisotropic heat kernels to enable higher expressivity. Monti \etal \cite{monti2017geometric} further improved the local geodesic disk convolution idea by introducing the notion of trainable local parameterization such that the network learns coordinate values of the neighboring points on geodesic disks from data. More recently, Honocka \etal \cite{hanocka2019meshcnn} similarly discretized tangent spaces on 1-ring neighborhoods of edges on a triangular mesh.

One problem is a majority of these works \cite{masci2015geodesic,masci2016geometric,monti2017geometric}, via the use of \textit{angular pooling} resolve the directional ambiguity caused by holonomy, end up supressing the features directionality. Recent works such as \cite{hanocka2019meshcnn} instead introduce the notion of invariant convolutions. Unfortunately, these methods achieve directional invariance at the cost of feature directionality giving them, in essence, the same issues. Verma \etal \cite{verma2018feastnet}, instead, proposed a data-driven scheme to attain directional correspondences among patches, by designing the network to learn the correspondence between the kernels. Some very recent works \cite{poulenard2018multi,schonsheck2018parallel,cohen2019gauge} contain a more mathematically rigorous solution to these directional ambiguities by introducing a vector field indicating the orientation of the local coordinate charts.

In summary, as opposed to the spectral methods, the spatial approaches do not assume global function bases, so that kernels are made explicit and compatible across different domains. Further, compare to spectral approaches, spatial methods tend to be computationally more efficient as they do not require an eigenvector computation. However, at the same time, mathematical rigor is often lost during the spatial discretization process. Furthermore, local neighborhood topology can vary across different locations on the manifold, rendering another obstacle for convolving kernels consistently.

The proposed method combines the advantages of spectral and spatial approaches. In our formulation, the convolution kernels are applied on local tensor fields spatially extracted from the surface. We parametrize the local tensor field via a set of corresponding coefficients of Zernike polynomials, which preserves the mathematical rigor similarly to spectral approaches.

\section{Zernike Convolutional Neural Networks}
In this section, we first introduce Zernike polynomials with their formal definition and examine their analytic properties. We employ them to describe the local geometry of a surface on the local tangent spaces. We will then define manifold convolution by using the notion of tangent spaces under Zernike formulation. Finally, we discretize the continuous formulation of the Zernike convolution to define ZerNet.

\subsection{Zernike polynomials}
Zernike polynomials ${Z}_i$ are an orthogonal polynomial basis for functions defined over the unit disk $\Omega \in \mathbb{R}^2$ such that $\langle {Z}_i, {Z}_j \rangle = \int_\Omega{{Z}_i(t){Z}_j(t)dt} = 0$ for all $i \neq j$ where $\langle \cdot, \cdot \rangle$ denotes the inner product. The formal definition of the Zernike polynomials is separated into even and odd sequences denoted in $Z_n^{m}(r,\theta)$ and $Z_n^{-m}(r,\theta)$ respectively:
\begin{align}
    \begin{split}
        {Z}_n^{m}(r,\theta) &= {R}_n^{m}(r) \cos(m \theta), \\
        {Z}_n^{-m}(r,\theta) &= {R}_n^{m}(r) \sin(m \theta),
    \end{split}
\label{eq:Zernike}
\end{align}
where $m$ and $n$ are non-negative integer indices with $m\leq n$; $r\in [0,1]$ is the radial distance; and $\theta$ is the azimuthal angle on the disk. Here, ${R}_n^{m}(r)$ is called the Zernike radial polynomial, defined as:
\begin{equation}
    R_n^{m}(r) = 
    \begin{cases}
    \mathlarger{\sum}_{k=0}^{\frac{n-m}{2}}
    {\frac{(-1)^{k}(n-k)!}{k!(\frac{n+m}{2}-k))!(\frac{n-m}{2}-k)!}r^{n-2k}}, & \text{if } n-m \text{ even}\\ \\
    0, & \text{otherwise}
    \end{cases}
    \label{eq:Zernike radial poly}
\end{equation}


\newcolumntype{P}[1]{>{\centering\arraybackslash}p{#1}}
\begin{figure}
\centering
\begin{tabular}{ P{2cm}  P{2cm}  P{2cm}  P{2cm}  P{2cm}  P{2cm} }
\includegraphics[width=0.08\textwidth]{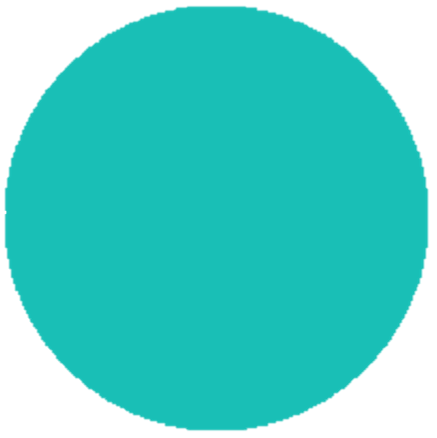} & \includegraphics[width=0.08\textwidth]{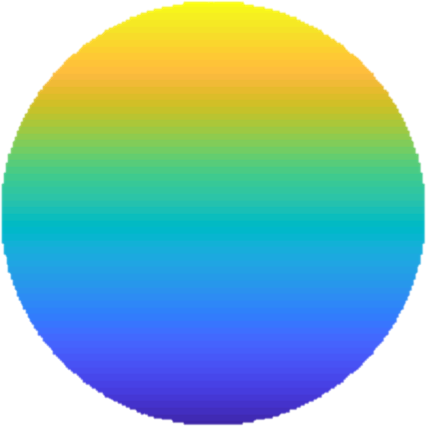} & \includegraphics[width=0.08\textwidth]{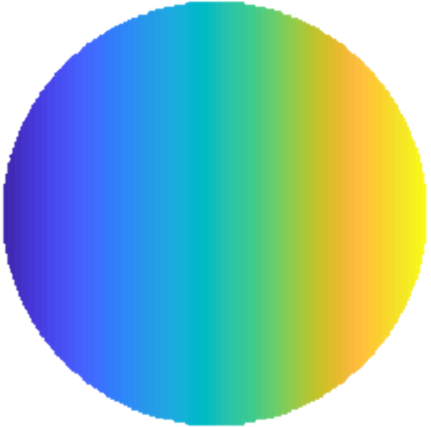} & \includegraphics[width=0.08\textwidth]{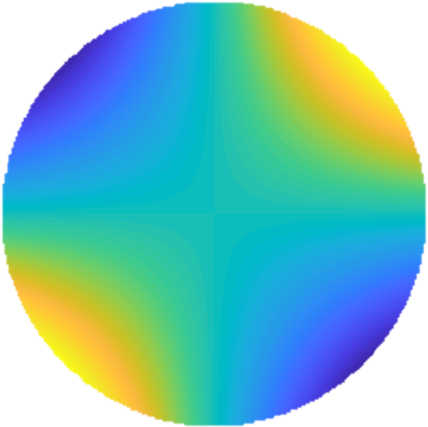} &
\includegraphics[width=0.08\textwidth]{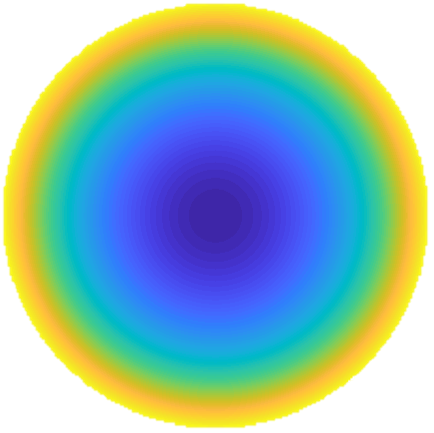} &
\includegraphics[width=0.08\textwidth]{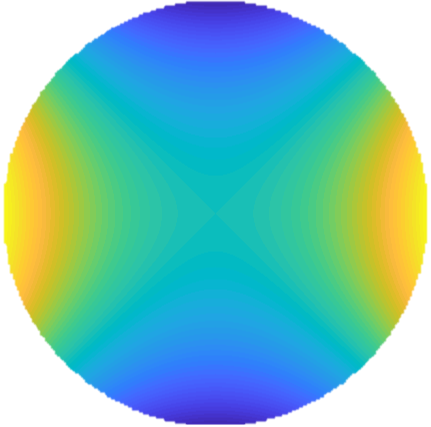}
\\
\small
$Z_0^{0}(Z_1)$ & $Z_1^{-1}(Z_2)$ & $Z_1^{1}(Z_3)$ & $Z_2^{-2}(Z_4)$ & $Z_2^{0}(Z_5)$ & $Z_2^{2}(Z_6)$
\\
\scriptsize Piston & \scriptsize Vertical tilt & \scriptsize Horizontal tilt & \scriptsize Oblique astigmatism & \scriptsize Defocus & \scriptsize Vertical astigmatism
\\
\includegraphics[width=0.08\textwidth]{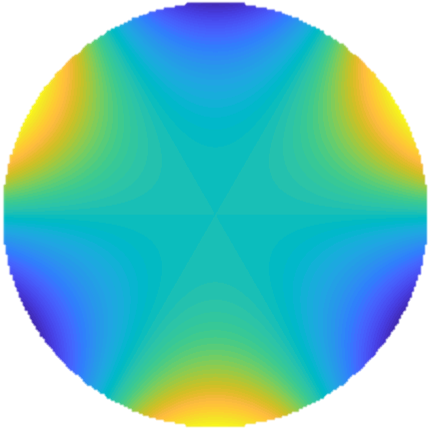} &
\includegraphics[width=0.08\textwidth]{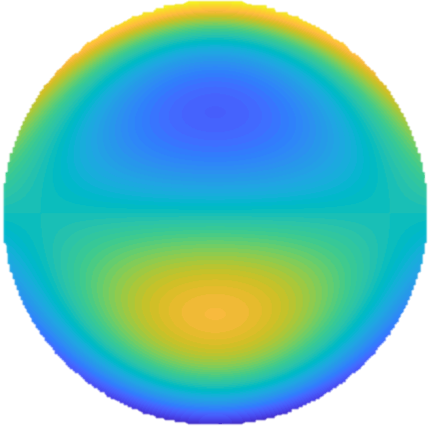} &
\includegraphics[width=0.08\textwidth]{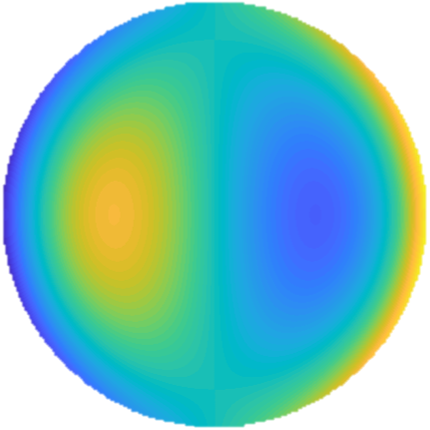} &
\includegraphics[width=0.08\textwidth]{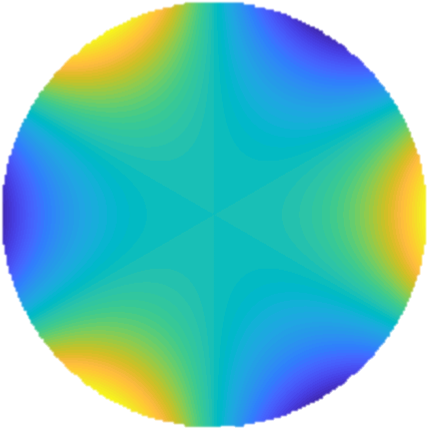} &
$\cdots$
\\
$Z_3^{-3}(Z_7)$ & $Z_3^{-1}(Z_8)$ & $Z_3^{1}(Z_9)$ & $Z_3^{3}(Z_{10})$ & $\cdots$
\\
\scriptsize Vertical trefoil & \scriptsize Vertical coma & \scriptsize Horizontal coma & \scriptsize Oblique trefoil
\end{tabular}
\caption{First few Zernike bases with optical interpretations.}
\label{fig:zernike_bases}
\end{figure}

Figure~\ref{fig:zernike_bases} illustrates the first few Zernike basis functions. Due to simple analytic properties, Zernike polynomials are widely used in optometry and ophthalmology to describe aberrations of the cornea or to represent lens distortions in optics. As we will see in later sections, we take advantage of such simple analytic properties of Zernike polynomial bases for extending CNNs on manifolds.

In practical use, Zernike polynomials can be normalized with a normalization factor $\sqrt{\frac{2-\delta(m,0)}{\pi}}$ such that the integral over the unit disk becomes unity:
\begin{equation}
    {{\hat{Z}_n^{m}}(r,\theta)} = Z_n^{m}(r,\theta) \sqrt{\frac{2-\delta(m,0)}{\pi}},
    \label{eq:noramlized Zernike func}
\end{equation}
where $\hat{Z}_n^{m}$ denotes the normalized Zernike polynomial and $\delta$ is the Kronecker delta function. Due to their orthonormality over the unit disk, the normalized Zernike polynomials can serve as bases for decomposing a complex function as a weighted sum, such that any function $f(r,\theta)$ defined on the domain $[0,1]\times[0,2\pi)$ can be expressed as:
\begin{equation}
    f(r,\theta) = \sum_{n=0}^{\infty}\sum_{m=-n}^{n}\alpha_{nm}{{\hat{Z}_n^{m}}(r,\theta)}.
    \label{eq:zernike_decomposition} 
\end{equation}

For the sake of brevity, in the rest of the paper we use the term
\emph{Zernike bases} standing for the normalized Zernike polynomials, denoted as
$Z_i(r, \theta)$ with index $i$ corresponding to a certain pair of $(n,m)$ in
(\ref{eq:zernike_decomposition}) following the ordering suggested by Figure
\ref{fig:zernike_bases}. Thus (\ref{eq:zernike_decomposition}) can be
expressed as:
\begin{equation}
    f(r,\theta) = \sum_{i=1}^{\infty}\alpha_{i}{Z_i(r,\theta)},
    \label{eq:zernike_decomposition_sim} 
\end{equation}
and we use the term \emph{Zernike coefficients} to represent the set of weights
$\alpha_{i}$ corresponding to \emph{Zernike bases} $Z_i$, which can be interpreted as
the coordinate vector of $f$ in the Zernike base space.

\subsubsection{Rotation}
\label{rotation property}
We want to be able rotate our functions in later sections, i.e, be able
to compute $f(r,\theta+\phi)$. To facilitate this, we establish some
rotational properties of Zernike polynomials. Using the sum of angle formulas for
trigonometric functions we obtain
\begin{align}
    \begin{split}
    Z_n^{m}(r,\theta+\phi) &= Z_n^{m}(r,\theta)cos(m \phi)-Z_n^{-m}(r,\theta)sin(m \phi)\\
    Z_n^{-m}(r,\theta+\phi) &= Z_n^{-m}(r,\theta)cos(m \phi)+Z_n^{m}(r,\theta)sin(m \phi)
    \end{split}
    \label{eq:Zernike_rotation}
\end{align}
Therefore, for a function $f$ decomposed as in
(\ref{eq:zernike_decomposition}) we then derive the following representation
of its rotation with an angle offset $\phi$ as:
\begin{align}
  \begin{split}
    f(r,\theta+\phi) &= \sum_{\text{even}}\alpha_{n}^{m} Z_n^m(r,\theta+\phi) + \sum_{\text{odd}}\alpha_{n}^{-m} Z_n^{-m}(r,\theta+\phi)\\
    &= \sum_{\text{even}}\tilde{\alpha }_{n}^{m} Z_n^m(r,\theta) + \sum_{\text{odd}}\tilde{ \alpha }_{n}^{-m} Z_n^{-m}(r,\theta),
  \end{split}
\end{align}
where $\tilde{\alpha}_i$ is computed via the rotational transform:
\begin{equation}
    \begin{bmatrix}
        \Tilde{\alpha}_{n}^{m} \\
        \Tilde{\alpha}_{n}^{-m}
    \end{bmatrix}
    = 
    \begin{bmatrix}
        cos(m\phi) & sin(m\phi)\\
        -sin(m\phi) & cos(m\phi)
    \end{bmatrix}
    \begin{bmatrix}
        \alpha_{n}^{m} \\
        \alpha_{n}^{-m}
    \end{bmatrix},
     \label{eq:rotational transform}
\end{equation}
To lessen the notational burden, for a function decomposed as in
(\ref{eq:zernike_decomposition_sim}), we will represent its rotation with an angle offset $\phi$ as:
\begin{equation}
  \label{eq:rotated_zernike_polynomial}
  (\rot(\phi))(f) = f(r,\theta+\phi) = \sum_{i=1}^{\infty}\tilde{\alpha}_i(\phi)Z_i(r,\theta),
\end{equation}
where $\tilde{\alpha}_i(\phi)$ corresponds to the rotated coefficient
from (\ref{eq:rotational transform}).
\begin{figure}[t]
\begin{center}
\includegraphics[width=0.65\linewidth]{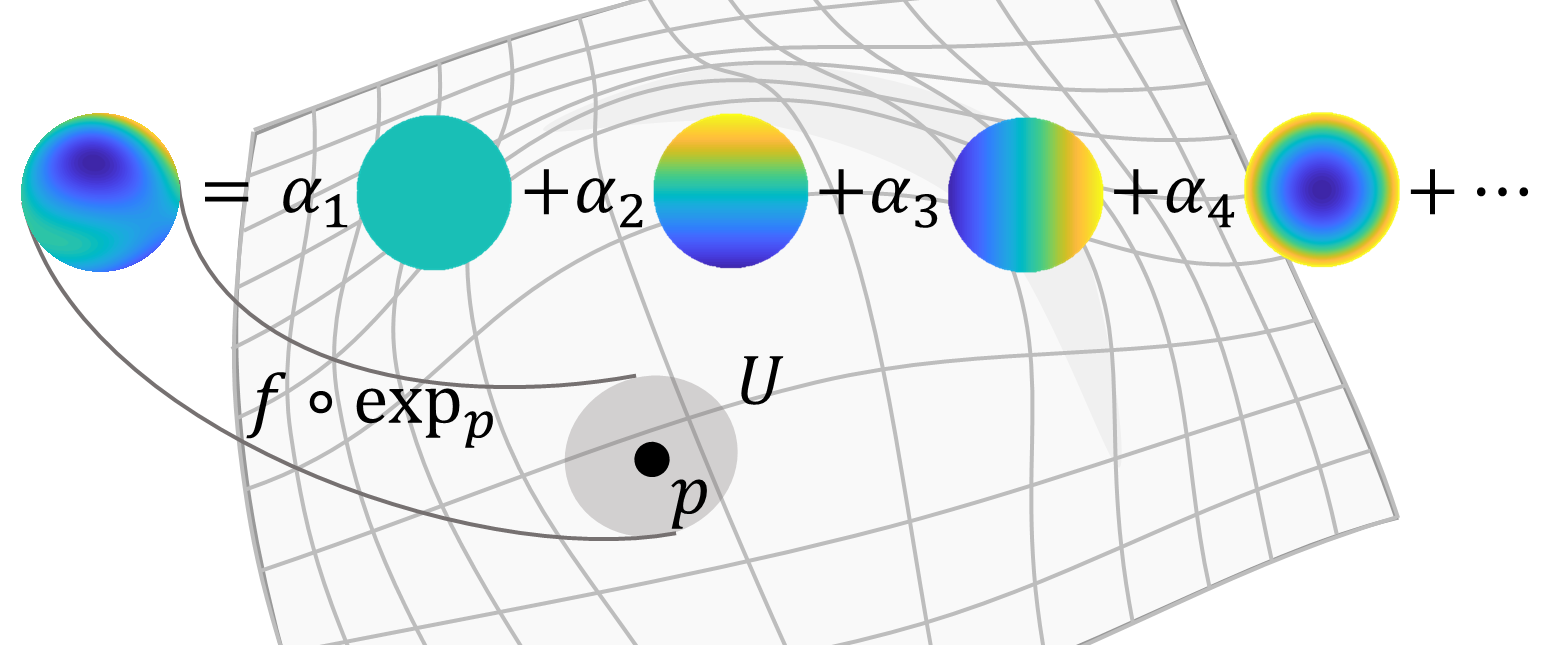}
\end{center}
    \caption{Zernike decomposition of a function.}
\label{fig:Zernike_decomposition}
\end{figure}

\subsection{Zernike convolution}
\label{Zernikeconvolution}
We generally view convolution on two-dimensional Cartesian space as the
``sliding'' of a filter over the image and measuring how much the image matches
our filter. Using this visual, we seek to define convolution on the manifold
similarly. Our filter will remain tangent to the manifold as we slide it around,
then we locally parameterize our space using the tangent space and do the usual
convolution. Thus we let $g: T_{p}\M \to \R$ be our ``filter'' and $f: \M \to
\R$ be our ``image''. The aim here is to show that if we represent
$f$ and $g$ locally at $p$ using Zernike polynomials as in
(\ref{eq:zernike_decomposition_sim}), the convolution on a manifold is nothing but a simple vector dot product of Zernike coefficient vectors on a tangent space and, thus, the convolution operation on a manifold is greatly simplified. We will also show that the rotation of a kernel also becomes a set of simple $2\times 2$ matrix rotations under this formulation. Hence, without compromising mathematical rigor, ZerNet can be defined on arbitrary surfaces efficiently while remaining compatible with existing tensor-based Euclidean CNN software packages.

\subsubsection{Convolution on manifold}
\label{sec: zernike_convolution}
\begin{wrapfigure}{r}{0.206\linewidth}
    \vspace{-0.7cm}
\centering
    \includegraphics[width=\linewidth]{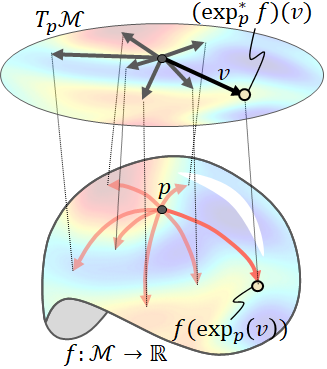}
    \vspace{-0.7cm}
\end{wrapfigure}
We would like to define the convolution on manifolds by using the notion of tangent spaces. First, for a given function $f:\M\rightarrow\R$ defined on a manifold $\M$, we can locally parameterize $f$ in terms of tangent vectors attached at $p\in\M$ by pulling back $f$ to the tangent space $T_p\M$ by the exponential map:
\begin{equation}
    {f}_p(v) \defas (\exp_p^*f)(v) = (f\circ\exp_p)(v),
    \label{eq:exponential_map}
\end{equation}
for all $v \in T_p\M$ and $\|v\| < R$, where $R$ defines the radius of the local neighborhood around $p$. The radius $R$ later will be weaved into the ZerNet implementation as a user-defined parameter concerning the size of the convolution kernel. Intuitively, we can think of (\ref{eq:exponential_map}) as projecting $v$ onto the manifold $\M$, sampling the function value of $f$ at the tip of the arrow, and bringing it back to $T_p\M$, for each and every $\|v\| < R$. 

\begin{wrapfigure}[10]{l}{0.206\linewidth}
    \vspace{-0.4cm}
\centering
    \includegraphics[width=\linewidth]{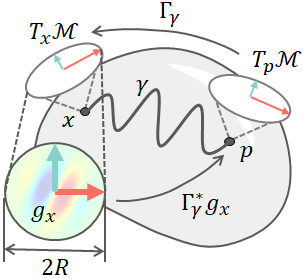}
    \vspace{-0.6cm}
\end{wrapfigure}
Now, let us consider a convolution kernel $g:\M\rightarrow\R$ defined around some ``origin'' $x\in\M$. Note that the origin $x$ here is a temporary reference we introduce to facilitate the discussion and will eventually disappear in the formulation. Hence, any arbitrary point on $\M$ can be chosen to serve as our $x$. As in (\ref{eq:exponential_map}), we can lift up the kernel $g$ to the tangent space $T_{x}\M$ by $g_{x}=g\circ\exp_{x}$. Under this setting, we can parallel transport $g_{x}$ from the tangent space at $x$ to the tangent space at $p$ along some path $\gamma \subset \M$ connecting $x$ and $p$. From the perspective of $p$, this is a process of pulling back the function $g_x$ from $T_x\M$ to $T_p\M$ (\ie coordinate transform from $T_x\M$ to $T_p\M$). Therefore, when the mapping $\Gamma_\gamma:T_p\M\rightarrow T_{x}\M$ is the isomorphism (\ie an invertible mapping) induced by the parallel transport along $\gamma$, the transported kernel $g_p: T_p\M \to \R^c$ is simply the pull-back $\Gamma_\gamma^*$ of the function $g_{x}$, or formally, $g_p = \Gamma_\gamma^* g_{x} = g_{x} \circ \Gamma_\gamma$. Under this setting, the convolution $f*g$ at $p$ is defined as:
\begin{equation}
    (f*g)(p) = \langle f_p, g_p \rangle = \int_{T_p\M}{f_p(v)g_p(v) dA}.
\end{equation}
Here, by representing $v$ in polar coordinates $v=(r, \theta)$ and by (\ref{eq:zernike_decomposition_sim}), we achieve
\begin{align}
    (f*g)(p) &= \iint_{r,\theta}{\sum_{i}\alpha_f^{i}{Z_i(r,\theta)}\sum_{i}\alpha_g^{i}{Z_i(r,\theta)} r\,dr\,d\theta} \nonumber \\
             &= \iint_{r,\theta}\sum_{i,j}\alpha_f^{i}\alpha_g^{j}{{Z_i(r,\theta)}{Z_j(r,\theta)} r\,dr\,d\theta}.
    \label{eq:zernike_local_integral}
\end{align}
Note, the area integral is about $r$ and $\theta$, but not the Zernike coefficients $\alpha^i$. Therefore, by the orthonormality $\int Z_i Z_j dA = \delta_i^j$, where $\delta_i^j$ is the Dirac delta function, the equation simplifies to:
\begin{equation}
    (f*g)(p) = \sum_i\alpha_f^{i}\alpha_g^{i},
    \label{eq:zernike_dot_product}
\end{equation}
which is merely a vector dot product between the Zernike coefficient vectors $\alpha_f$ and $\alpha_g$ at $p$.

Furthermore, if we assume the isomorphism $\Gamma_\gamma$ is isometry between $T_{p}\M$ and $T_{x}\M$, $g_x$ and $g_p$ are the same up to rotation. Therefore, in practical use, we do not need to actually do anything to pull back $g_x$ to $T_p\M$ but, instead, just reuse the same $r$, $\theta$ representation of $g_x$ for $g_p$ with an angle offset $\phi$, such that $g_p = (\rot(\phi))(g_x)$. Meanwhile, from the proof given in Section~\ref{rotation property}, the rotational transformation $\rot(\phi)$ of $g_x$ can be achieved simply as a set of $2 \times 2$ matrix rotations of the Zernike coefficient vectors, presenting an analytic, vector-space formula for the parallel transport.

\subsubsection{Discretization}
\label{sec: Discretization}
With the above continuous theory, discretization is rather straightforward. Here, with a remark that the above continuous theory can be discretized to virtually any type of discretization, be it point clouds, polygonal meshes, and parametric surfaces, we demonstrate here only the triangular mesh case for brevity.

\begin{wrapfigure}[7]{r}{0.206\linewidth}
    \vspace{-0.7cm}
\centering
    \includegraphics[width=\linewidth]{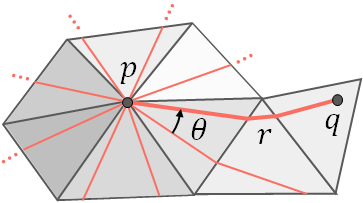}
    \vspace{-0.6cm}
\end{wrapfigure}
\noindent \textbf{Exponential map} \indent Computation of a discrete exponential map has been widely studied in computer graphics and relevant areas, for a variety of purposes such as interactive drawing on surfaces \cite{schmidt2006interactive} and local remeshing \cite{melvaer2012geodesic}. These methods aim to find the shortest geodesic paths from a reference point to its neighbors using variants of the Dijkstra's algorithm. With the geodesic distance, we can then determine the radius $r$. Furthermore, by measuring the angle from some reference direction to the geodesic curve, we can also determine the azimuth $\theta$. Here, the reference direction typically is arbitrarily picked, most commonly based on the local ordering (indexing) of the neighboring points. To this end, algorithms such as \cite{melvaer2012geodesic} can be applied to triangular and more general polygonal meshes, while Crane \etal~\cite{Crane:2017:HMD} can be extended to a much broader scope including point clouds, subdivision surfaces, noisy/partial meshes, and spline surfaces.

\noindent \textbf{Zernike decomposition} \indent From (\ref{eq:zernike_decomposition_sim}) and the \textit{geodesic polar coordinates} $(r_j, \theta_j)$ computed via the discrete exponential map, the relationship $f_p(r_j, \theta_j) = \sum_{i}{\alpha_f^{i} Z_i(r_j, \theta_j)}$ holds true for all $(r_j, \theta_j)$ in the neighborhood. This in fact is a linear system of equations solved locally at $p$. In implementation, the summation is approximated with a finite number of terms $k$ instead of the infinite sum. The number of points in the neighborhood may be lesser than $k$. In this case, we simply sample more points by linearly blending the existing points and their function values. Practically, we sample a greater number of points in the neighborhood than $k$ and solve the linear system in a least square manner.

\noindent \textbf{Convolution} \indent With the above discretization, the local parameterization $f_p$ can be represented as a $k \times d$ tensor, where $k$ is the number of Zernike polynomials used in decomposition and $d$ is the number of image channels. If a manifold is discretized with $N$ points, the stack $F=[f_p]$ of parameterized functions is an $N \times k \times d$ tensor. Meanwhile, the convolution kernel $g$ can also be similarly discretized as $k \times d$. However, due to the rotation of the kernel induced by parallel transport, the tensor $G$ corresponding to $g$ requires an additional axis for the rotation parameter, such that $G$ becomes $k \times d \times s$, where $s$ is the \textit{angular resolution}. Finally, from the fact that Zernike convolution is a simple dot product between the coefficient vectors (\textit{i.e.,} (\ref{eq:zernike_dot_product})), the implementation of the Zernike convolution layer becomes a simple tensor dot product between $F$ and $G$ along $k$ and $d$ axes, producing an $N \times s$ response.

\noindent \textbf{Angular pooling} \indent With the response produced via convolution, the activation maps defined across different direction configurations along the angular axis can be down-sampled via angular-pooling operation. Similar to the (spatial) max-pooling layers in conventional CNNs, angular-pooling selects the maximum value among the activations produced by convolution kernels in different orientations, producing $N$ response.
\section{Experiments}
We validate the proposed approach against two supervised learning tasks, namely, classification and regression. For classification, we test our algorithm against two popular problems in geometry processing: point-wise correspondence matching and semantic segmentation, which are both commonly formalized as per-vertex classification problems. For regression, we introduce a new data set in which the goal is to predict a scalar field defined on a surface.

\subsection{3D shape correspondence}
\label{Faust exp}
The goal of 3D shape correspondence problem is to find semantically meaningful one-to-one matching between points on a query surface and points on a reference surface.

\noindent \textbf{Data set} \indent We employ the FAUST human data set \cite{Bogo:CVPR:2014} with the similar experimental setup as in other state-of-the-art methods \cite{masci2015geodesic,boscaini2016anisotropic,monti2017geometric,verma2018feastnet}. The data set consists of $100$ watertight meshes of $10$ different subjects with $10$ different poses for each. Each mesh contains $6,890$ vertices and the semantic correspondence among the vertices are already established in the data set. We utilize this as a ground truth for the training, but pretend no such information is provided in testing.

\noindent \textbf{Input processing} \indent We normalize the 100 mesh models in FAUST data set to have the same surface area, $15,000 \; cm^2$ with an approximate shape diameter of $200 \; cm$. For initial input, we take the canonical XYZ-coordinates of mesh vertices as the input to the network.

\noindent \textbf{Network architecture and parameter setting} \indent
For fair comparison with the other state-of-the-art methods \cite{masci2015geodesic,boscaini2016anisotropic,monti2017geometric,verma2018feastnet}, we first test $\emph{Conv64} \rightarrow \emph{Conv128} \rightarrow \emph{Conv256} \rightarrow \emph{Lin512} \rightarrow \emph{Lin6890+softmax}$, a single-scale architecture that generalizes the other methods. The numbers in the layer names indicate the dimension of the output channel. In each of the \emph{Conv} block, we set the kernel size $r_0 = 5.5 cm$ for computing the local exponential map. For the discretization of local exponential map, we first uniformly sample 12,000 surface points over the entire surface, and collect 50 sampled points in the neighborhood of every mesh vertex within the radius $r_0$. The first 21 \emph{Zernike bases} ($k=21$) are used for Zernike decomposition.

\begin{figure}[H]
\begin{center}
\includegraphics[width=0.65\linewidth]{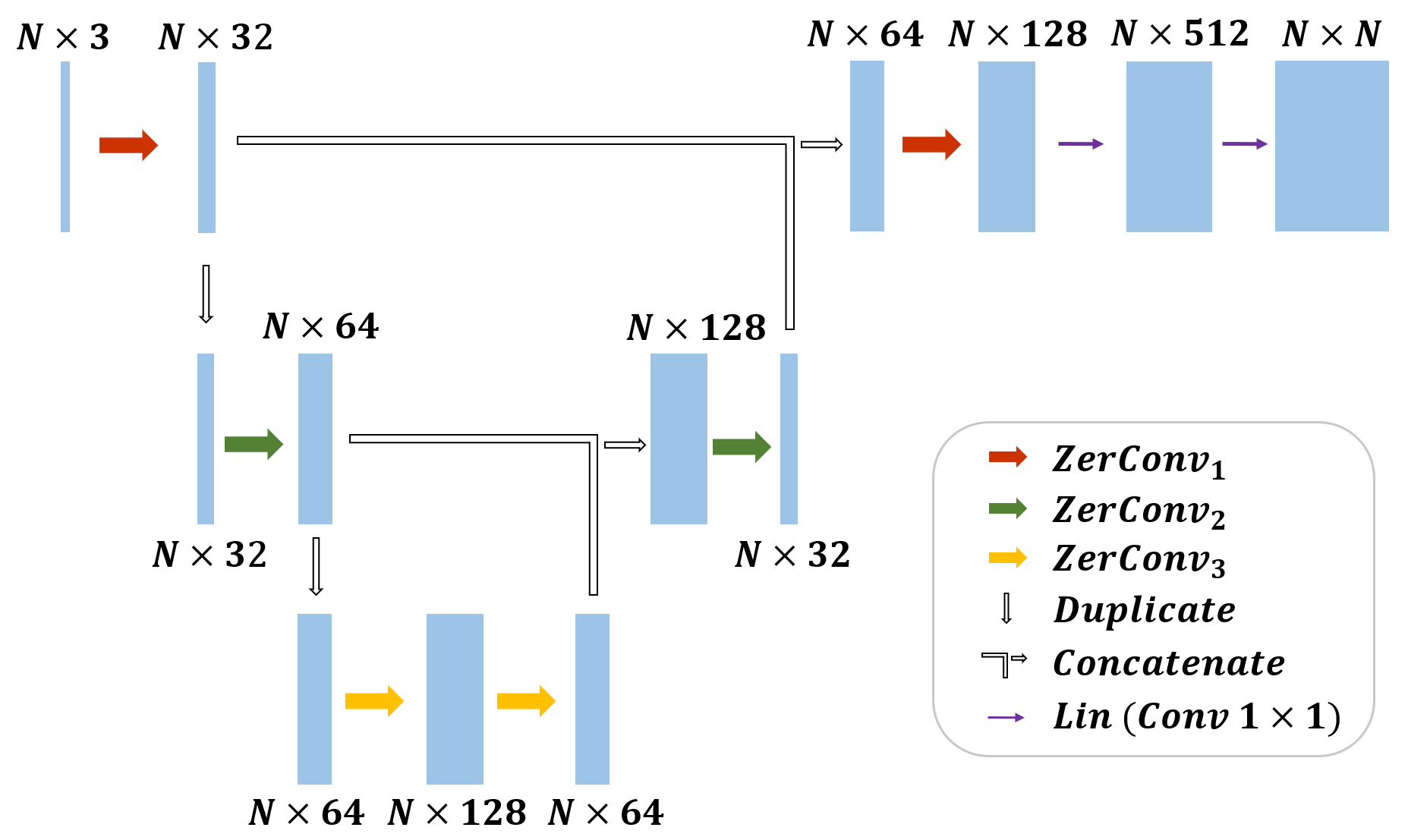}
\end{center}
  \caption{Our multi-scale ZerNet architecture}
\label{fig:multiscale_ZerNet}
\end{figure}

Meanwhile, a multi-scale architecture to incorporate higher contextual information is proposed in \cite{verma2018feastnet}, and we also compare our method in the multi-scale setting. To this end, we set the kernel sizes ($r_0$) as 4.5, 5.5 and 6.75 cm respectively for three scales of \emph{Conv} blocks ($ZerConv_1$, $ZerConv_2$ and $ZerConv_3$ as in Figure \ref{fig:multiscale_ZerNet}). For discretization, we uniformly sample 18,000, 12,000 and 8,000 surface points over the mesh surface, respectively, for the three scales of \emph{Conv} blocks. Note that the total number of sample points over the surface is inversely proportional to the square of $r_0$. Thus, for different scales, the number of discretized samples surrounding the mesh vertex within its local exponential map remains approximately the same. We collect 50 discretized samples in the neighborhood and use the first 21 \emph{Zernike bases} ($k=21$) for Zernike decomposition.

The above two models were trained using the Adam optimizer \cite{DBLP:journals/corr/KingmaB14} with the sparse categorical cross-entropy loss as the objective.

\begin{figure}[H]
\begin{center}
\includegraphics[width=0.55\linewidth]{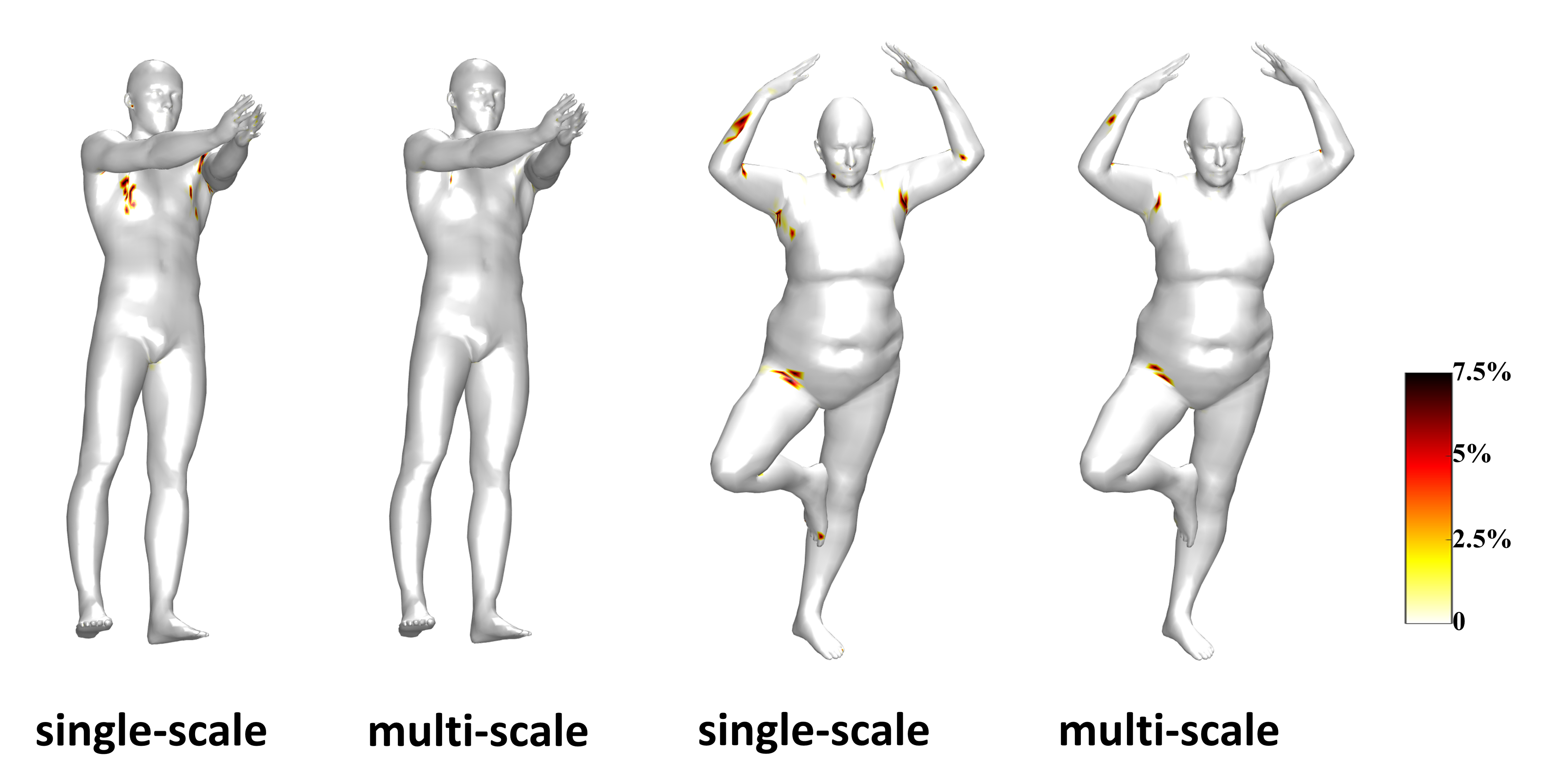}
\end{center}
  \caption{Geodesic errors (in \% of the shape diameter) on two test shapes estimated using our single-scale and multi-scale ZerNet architectures}
\label{fig:ZerNet_single_Vs_multi}
\end{figure}

\begin{figure*}[ht]
    \centering
    \begin{subfigure}[b]{\linewidth}
        \centering
        \includegraphics[width=\linewidth]{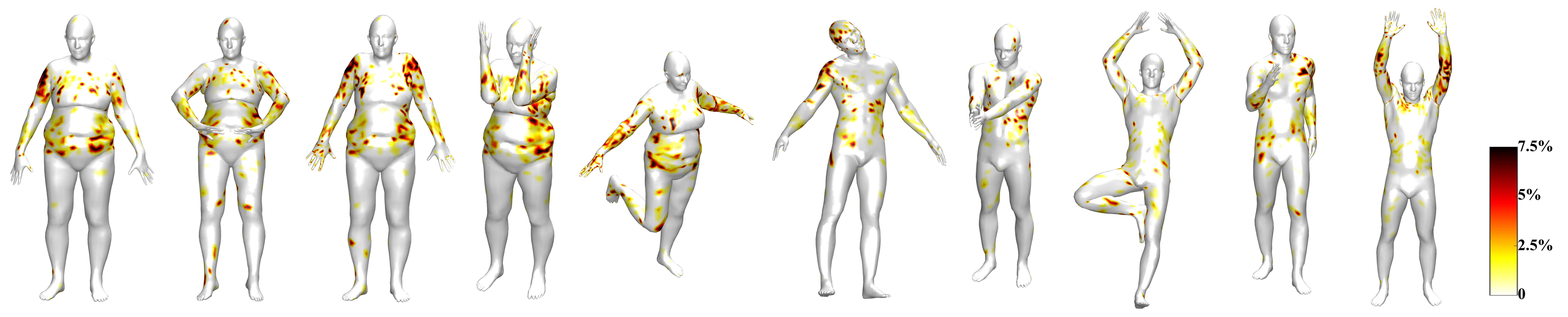}
        \caption{FeaStNet}
        \label{fig: FeaStNet geodesic error}
    \end{subfigure}
    \begin{subfigure}[b]{\linewidth}
        \centering
        \includegraphics[width=\linewidth]{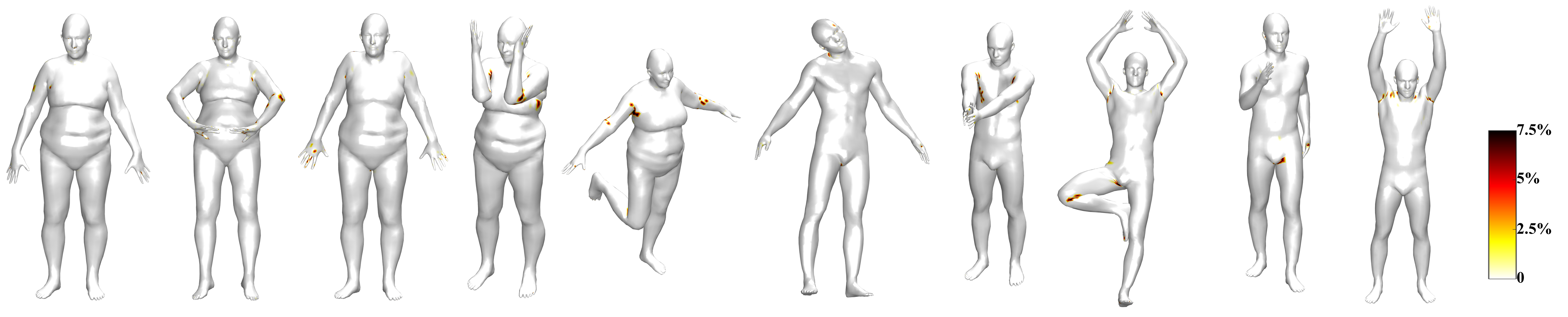}
        \caption{ZerNet}
        \label{fig: ZerNet geodesic error}
    \end{subfigure}
    \caption{Point-wise geodesic error (in \% of the shape diameter) of ZerNet vs FeaStNet on the FAUST human dataset \cite{Bogo:CVPR:2014}. Results are generated based on raw performance of both approaches shared a similar single-scale architecture.}
    \label{fig: Faust geodesic error}
\end{figure*}
\noindent \textbf{Results} \indent
To evaluate the performance, we first compare the classification accuracy, defined as:
\begin{equation}
    \text{Accuracy} = \frac{\text{True Classification}}{\text{Total Number of Surface Points}}.
\end{equation}
The test result is reported in Table \ref{tbl:faust_accuracy}. As shown in the table, with single-scale architecture, ZerNet outperformed all other state-of-the-art methods with the accuracy of 94.7\%, even without additional refinement post-processes \cite{Ovsjanikov:2012,vestner2017product}. In multi-scale setting, FeaStNet \cite{verma2018feastnet} outperformed ZerNet with less than 2\% margin.

\begin{figure}[H]
\begin{center}
\includegraphics[width=0.55\linewidth]{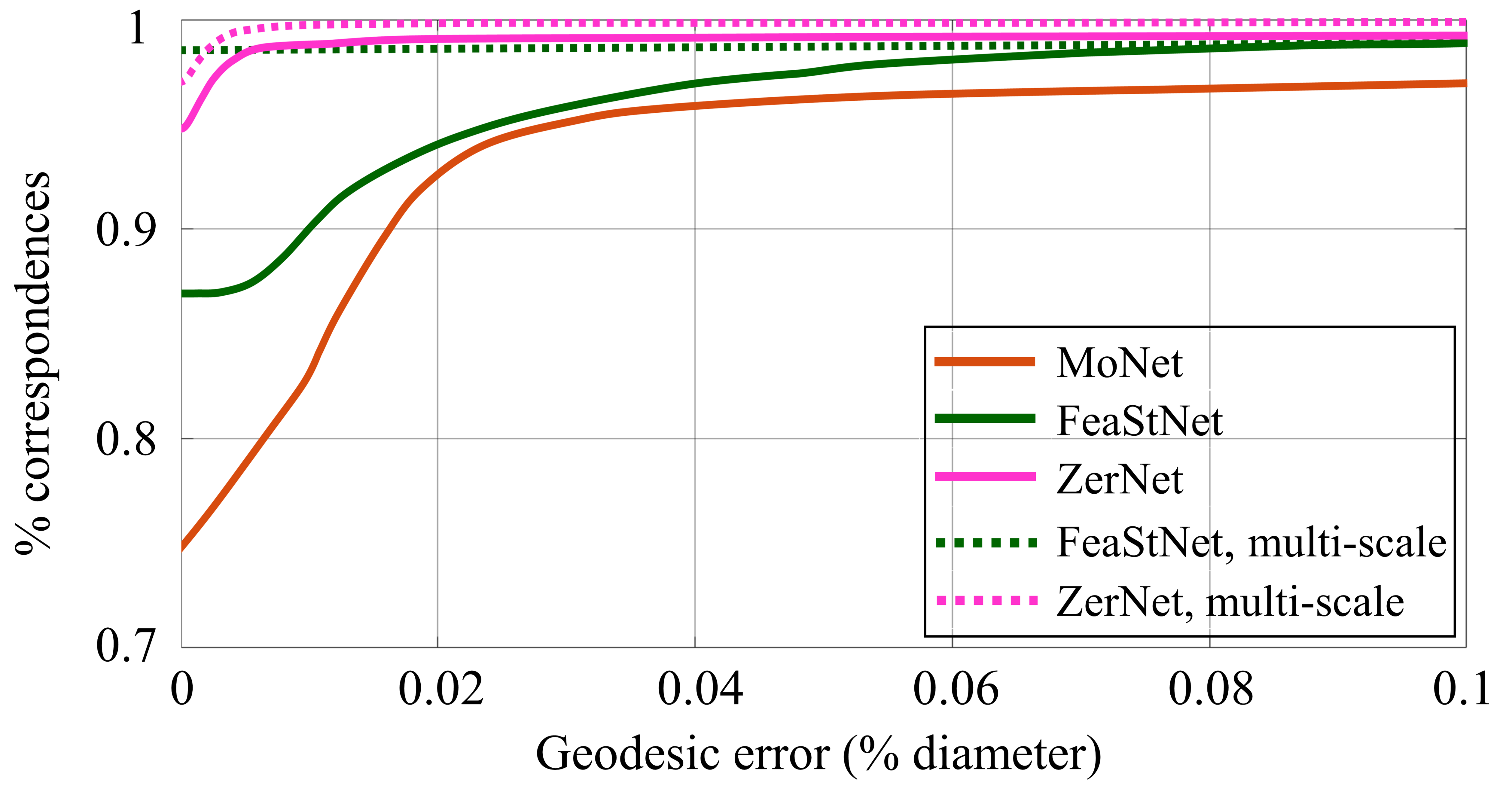}
\end{center}
   \caption{Shape correspondence quality obtained by different methods on the FAUST human data set. Results are generated based on the raw performance of each method without additional post-process refinements.}
\label{fig:fault_quality}
\end{figure}

We further evaluated the quality of correspondence using the Princeton Benchmark \cite{kim2011blended}. The quality of correspondence is measured by the percentage of correctly predicted matches within a geodesic disk around the ground-truth. The results were plotted in Figure \ref{fig:fault_quality} with varying radii of the geodesic disk from 0\% to 10\% of shape diameter. In single-scale, our method demonstrates a significantly better quality of correspondence than all the other benchmark methods. In multi-scale, FeaStNet shows the top performance at the zero radius (smallest error tolerance). With larger radii, however, ZerNet begins to outperform FeaStNet.

\begin{table}[h]
\begin{center}
\begin{tabular}{lrr}
\hline
    Method & Input & Accuracy \\
\hline
    ACNN \cite{boscaini2016anisotropic} w/o refinement & SHOT & 60.6\% \\
    ACNN \cite{boscaini2016anisotropic} w/ refinement \cite{Ovsjanikov:2012} & SHOT & 62.4\% \\
    GCNN \cite{masci2015geodesic} w/o refinement & SHOT & 65.4\% \\
    GCNN \cite{masci2015geodesic} w/ refinement \cite{Ovsjanikov:2012} & SHOT & 42.3\% \\
    MoNet \cite{monti2017geometric} w/o refinement & SHOT & 73.8\% \\
    MoNet \cite{monti2017geometric} w/ refinement \cite{vestner2017product} & SHOT & 88.2\% \\
    FeaStNet \cite{verma2018feastnet} w/o refinement & XYZ & 88.1\% \\
    FeaStNet \cite{verma2018feastnet} w/ refinement \cite{vestner2017product} & XYZ & 92.2\% \\
    ZerNet (Ours) w/o refinement & XYZ & \textbf{94.7}\%\\
\hline
    FeaStNet \cite{verma2018feastnet} multi-scale & XYZ & \textbf{98.7}\% \\
    ZerNet (Ours) multi-scale & XYZ & 96.9\%\\
\hline
\end{tabular}
\end{center}
\caption{Correspondence accuracy on the FAUST human data set of our method and recent state-of-the-art manifold convolution approaches. Accuracies for the compared methods \cite{masci2015geodesic,boscaini2016anisotropic,monti2017geometric,verma2018feastnet} are directly taken from the corresponding papers.}
\label{tbl:faust_accuracy}
\end{table}

Figure \ref{fig:ZerNet_single_Vs_multi} visualizes the point-wise geodesic correspondence errors on two representative shapes of the two test subjects using our single-scale and multi-scale ZerNet architectures. Figure \ref{fig: Faust geodesic error} shows the point-wise geodesic correspondence error of our method in comparison with the most recent state-of-the-art method, FeaStNet \cite{verma2018feastnet}, based on the similar single-scale architecture.

\subsection{Semantic segmentation}
\label{Segmentation exp}
In addition, we validate ZerNet on semantic segmentation problem. We compared ZerNet against other state-of-the-art methods, including Toric-cover CNN \cite{maron2017convolutional}, PointNet++ \cite{qi2017pointnet++}, Dynamic graph CNN \cite{wang2018dynamic} and MDGCNN \cite{poulenard2018multi}.

\noindent \textbf{Data set} \indent 
We use the human segmentation benchmark in \cite{maron2017convolutional} for comparison. The training set consists of 370 models collected from SCAPE, FAUST, MIT and Adobe Fuse \cite{Adobe2016}. All models are manually segmented into eight labels, one for the head, one for the torso, three for the arms and three for the legs. The test set is the 18 models collected from the SHREC07 data set in ``human'' category.

\noindent \textbf{Experiment setting} \indent
We normalize all human models to have the same surface area, $15,000 \; cm^2$, and take XYZ-coordinate of mesh vertices as the input to ZerNet. For the network architecture and parameters setting, we follow the same setting as the single-scale architecture used for the shape correspondence experiment (Section~\ref{Faust exp}). Compared to the dense correspondence task, as the classification classes required for segmentation is significantly reduced ($6,890 \rightarrow 8$), a similar architecture but with less output channel dimensions of each layer is adopted: $\emph{Conv32} \rightarrow \emph{Conv64} \rightarrow \emph{Conv128} \rightarrow \emph{Lin256} \rightarrow \emph{Lin8+softmax}$.

\begin{figure}[h]
\begin{center}
\includegraphics[width=0.55\linewidth]{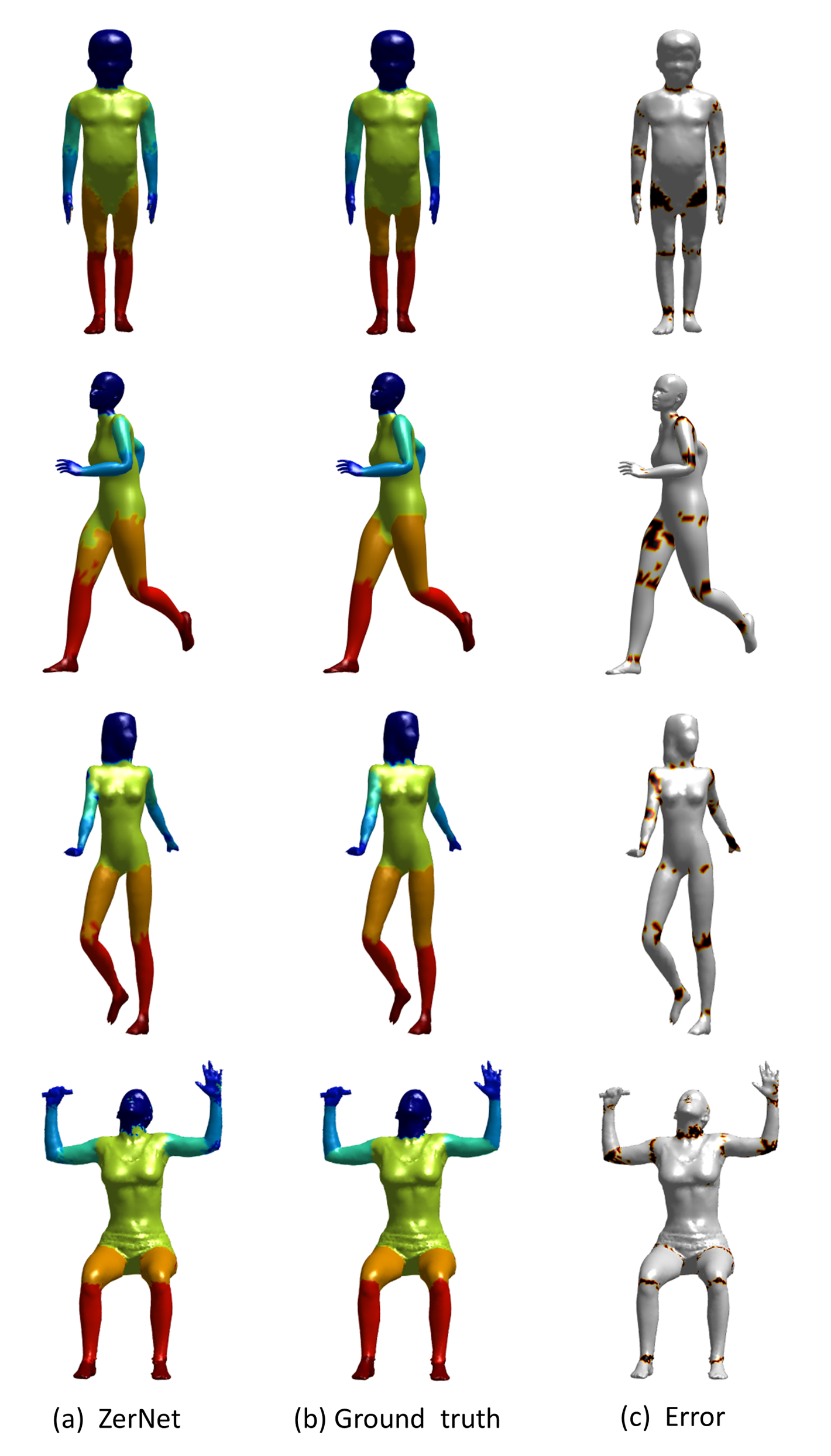}
\end{center}
   \caption{Semantic segmentation on human body shapes via ZerNet compared with ground truth. Miss segmentation regions on third column.}
\label{fig:Seg_vis}
\end{figure}

\noindent \textbf{Results} \indent 
The segmentation accuracy is reported in Table~\ref{tbl:seg_accuracy}. As shown in the table, ZerNet achieves a high accuracy comparable to the other state-of-the-art methods. As shown in Figure~\ref{fig:Seg_vis} (a), the overall segmentation quality of ZerNet was satisfactory, except at the segment boundary. Some of those errors (\eg right thigh in the second row of Figure~\ref{fig:Seg_vis}) were clearly misclassifications. However, in many cases, we noticed the error was due to the variations and inconsistencies in human-labeled ground truth.

\begin{table}[H]
\begin{center}
\begin{tabular}{lrr}
\hline
    Method & Input & Accuracy \\
\hline
    Toric cover \cite{maron2017convolutional} & WKS,AGD,curv. & 88\% \\
    Pointnet++ \cite{qi2017pointnet++} & XYZ & \textbf{90.8}\% \\
    DynGraphCNN \cite{wang2018dynamic} & XYZ & 89.7\% \\
    MDGCNN \cite{poulenard2018multi} & XYZ & 88.6\% \\
\hline
    ZerNet (Ours) & XYZ & \textbf{88.7}\%\\
\hline
\end{tabular}
\end{center}
\caption{Segmentation accuracy on the human body data set introduced in \cite{maron2017convolutional} of our method and several state-of-the-art methods. Accuracies for the compared methods \cite{maron2017convolutional,qi2017pointnet++,wang2018dynamic,poulenard2018multi} are directly taken from \cite{poulenard2018multi}.}
\label{tbl:seg_accuracy}
\end{table}

\subsection{Aneurysm wall stress estimation}
\label{Aneurysm exp}
Lastly, we validate our method on a scalar field regression task. Specifically, the problem is to estimate the mechanical stress distributed over the surface of cerebral aneurysm.

\noindent \textbf{Data set} \indent 
To this end, we introduce a new benchmark data set comprised of 3D surface meshes of 26 cerebral aneurysm cases. According to relevant literature \cite{lu2007inverse,lu2008inverse,lu2016solving,luo2018machine}, the magnitude of the mechanical stress distribution on aneurysm is known to be correlated with the local surface geometry. The goal here, therefore, is to utilize CNNs to predict the stress distribution on aneurysm based on surface geometry. This is essentially a scalar-field regression problem defined on a surface. The ground-truth values are computed from finite-element (FE) simulations. The aneurysm models are different in mesh topology such that the number of vertices and how the vertices are connected are inconsistent across different models. The total number of mesh vertices across the aneurysm models varies from 1,135 to 8,197 and the surface area is in a range between $27.72$ and $169.23$.

\noindent \textbf{Input processing} \indent
We first normalize the 26 aneurysm models to have the same surface area of 100. Wall stress distribution is also scaled accordingly based on the fact that the wall stress is proportional to the square root of surface area as the governing physics equation states. We then uniformly sample $8,000$ random points on each mesh surface alongside the stress value, with an assumption that the stress distribution is piece-wise linear at each triangle.

\noindent \textbf{Network architecture and parameter setting} \indent
We again use the similar architecture as in the single-scale shape correspondence experiment: $\emph{Conv128} \rightarrow \emph{Conv256} \rightarrow \emph{Conv512} \rightarrow \emph{Lin800} \rightarrow \emph{Lin1}$. We set $r_0 = 0.6$ for computing the local exponential map. We use the first 21 \emph{Zernike bases} ($k=21$) for Zernike decomposition. The Adam optimizer was used for training, with the mean squared error (MSE) loss between ground truth and stress prediction on all sampled points of each aneurysm as the objective.

\noindent \textbf{Results} \indent 
We cross-validated our method along with the current state-of-the-art FeaStNet \cite{verma2018feastnet}. Among the 26 aneurysm surfaces, we randomly selected five cases for performing leave-one-out cross validations. For each of the cases among the five, we left it out as the validation set and used the rest 25 models in our data set as the training set to train the ZerNet or FeaStNet.

\begin{figure}[H]
\begin{center}
\includegraphics[width=0.9\linewidth]{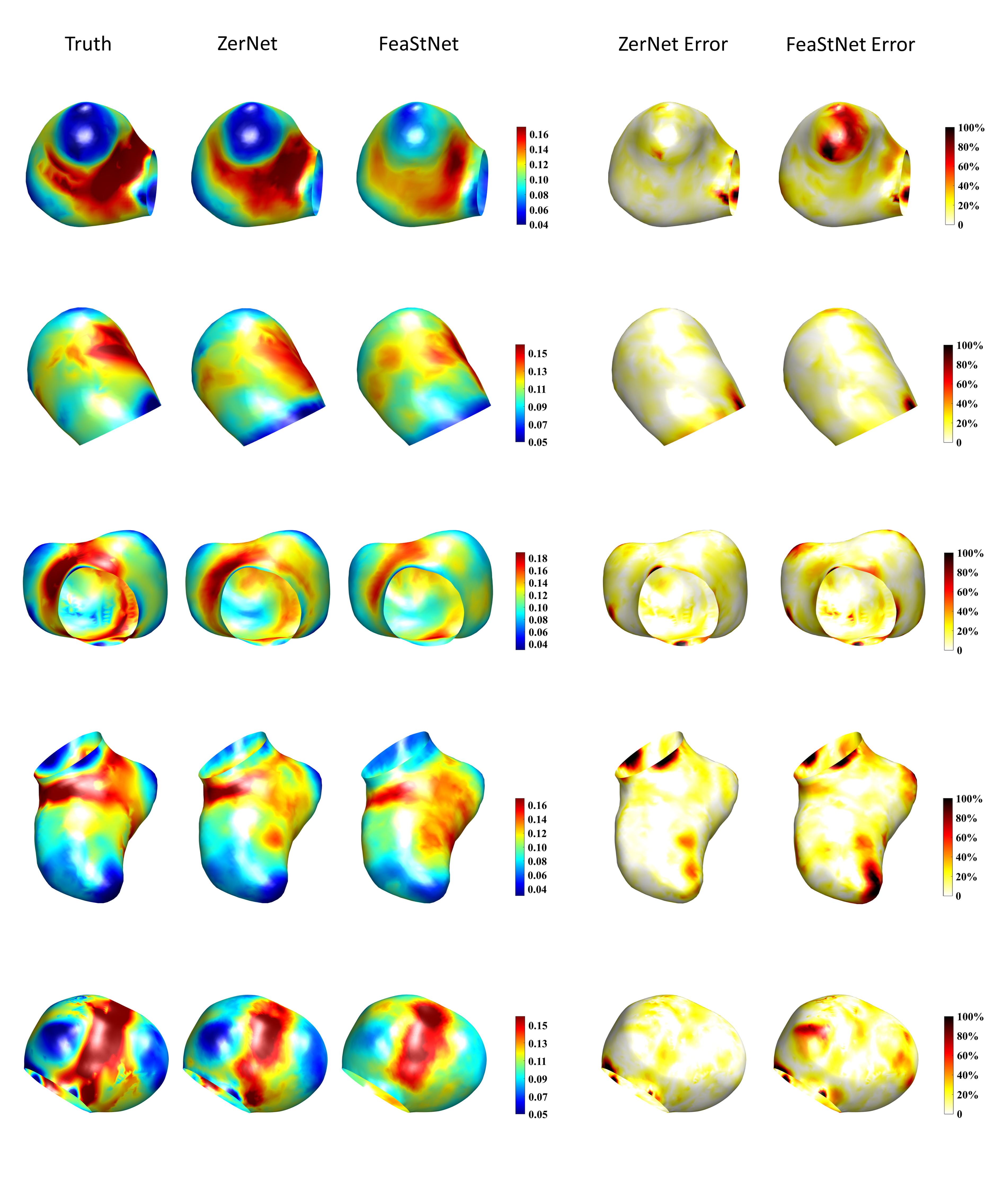}
\end{center}
   \caption{Visualization of the estimated wall stress distribution and the distributed absolute percentage errors on five validation cases. Each row corresponds to the row of the same order in Table~\ref{Prediction-table}.}
\label{fig:aneurysm color mapping}
\end{figure}

The result is reported in Table~\ref{Prediction-table}. In addition, Figure~\ref{fig:aneurysm color mapping} visualizes the stress estimation results on the five validation cases and their point-wise error (in percentage of the true stress value on each vertex). We can notice that ZerNet outperformed FeaStNet for all metrics as presented in Table~\ref{Prediction-table}, and achieved a satisfactory quality of stress estimation significantly better than FeaStNet.

\begin{table*}[ht]
  \centering
  \smaller
  \begin{tabularx}{\linewidth}{l|rrrrrr|rrrrrr}
    \toprule
     & \multicolumn{6}{c|}{ZerNet} & \multicolumn{6}{c}{FeaStNet} \\
    Model ID & MAPE & RRMSE & PCC & HR$_{10}$ & HR$_{20}$ & HR$_{30}$ & MAPE & RRMSE & PCC & HR$_{10}$ & HR$_{20}$ & HR$_{30}$\\
    \midrule
    TPIa105I & 10.11\% & 13.04\% & 0.91 & 65.84\% & 90.81\% & 97.24\% & 16.09\% & 19.98\% & 0.80 & 48.66\% & 74.37\% & 86.36\%\\
    TPIa166I & 9.57\% & 12.97\% & 0.87 & 68.92\% & 90.75\% & 95.21\% & 13.38\% & 15.86\% & 0.79 & 45.72\% & 79.79\% & 92.89\%\\
    TPIa182I & 14.32\% & 18.70\% & 0.87 & 45.63\% & 77.60\% & 92.58\% & 19.99\% & 24.60\% & 0.76 & 35.71\% & 64.03\% & 79.27\%\\
    TPIa32I  & 13.80\% & 16.32\% & 0.88 & 50.48\% & 82.53\% & 93.24\% & 23.89\% & 25.73\% & 0.65 & 28.37\% & 54.48\% & 73.38\%\\
    TPIa33I  & 9.60\%  & 13.98\% & 0.90  & 65.30\% & 90.70\% & 97.61\% & 13.78\% & 17.71\% & 0.82 & 52.25\% & 78.92\% & 89.92\%\\
    \bottomrule
  \end{tabularx}
  \caption{Point-wise regression result of ZerNet vs FeaStNet over five validation cases. Performance was measured with the following criteria: mean absolute percentage error (MAPE), relative root mean square error (RRMSE), Pearson correlation coefficient (PCC) and hit-rate (HR). Subscripts under HR represents the tolerance threshold (10\%, 20\% and 30\%) and HR was calculated as the percentage ratio of the number of the vertices that have the scalar values accurately predicted over the total number of mesh vertices.}
  \label{Prediction-table}
\end{table*}

It should be noted that neither ZerNet nor FeaStNet is successful in predicting correct values near the boundary. This may be due to incorrect boundary conditions (``zero-padding'' is enforced currently). Hence, it would be worthwhile to investigate practical ways to assign boundary conditions in future research.

\subsection{Directional Convolution}
\label{sec:discussion}
As discussed in Section~\ref{relatedworks}, a critical bottleneck in many geometric CNNs is the suppression of feature directionalities, resulting from the path-dependency of the kernel rotation during parallel transport. Under the current setting of ZerNet, we apply angular pooling after each convolution. However, directional features would be neglected under this setting and the angular pooling surpresses the directionality. Similar to the idea proposed in \cite{poulenard2018multi}, we can re-define a \textit{directional function} $f_p(r, \theta; \phi)$ on each tangent space $T_p\M$, as a function $f_p(r,\theta)$ parameterized by an additional direction parameter $\phi$. Then, the activation maps in convolutional neural networks are modeled as directional functions to resolve ambiguities in feature directions. The advantage of ZerNet, \textit{i.e.,} the fact that a rotation of a function is simply a $2\times 2$ matrix rotation, allows us to write an analytic formula for directional functions, as opposed to discrete angular bins used in \cite{poulenard2018multi}, presenting a straightforward implementation. 

The single-scale correspondence matching experiment in Section~\ref{Faust exp} and the wall stress estimation experiment in Section~\ref{Aneurysm exp} have been re-done with the new direction-preserving convolution. For both experiments, we set the angular resolution $s=4$ and preserve the direction axis in each activation map between \emph{Conv} blocks. For the correspondence matching experiment, we observe a noticeable increase in accuracy from 94.7\% to 96.1\%. For the wall stress estimation experiment, we also observed meaningful improvement in prediction accruacy as in Table~\ref{New_Prediction_table}. We also have tested in higher angular resolutions \ie{$s=8, 16$}, but the improvement of the performance was limited, whereas the computational time increased dramatically (Table~\ref{computing_time_table}). This awaits further investigation and optimization.

\begin{table}[h]
  \centering
  \begin{tabularx}{0.55\linewidth}{l |r r r r r}
    \toprule
     & \multicolumn{5}{c}{ZerNet w/ direction preserving} \\
    Model ID & MAPE & RRMSE & PCC & HR$_{10}$ & HR$_{20}$  \\
    \midrule
    TPIa105I & 9.97\% & 13.15\% & 0.91 & 68.64\% & 91.31\%\\
    TPIa166I & 9.90\% & 12.87\% & 0.87 & 65.33\%  & 91.18\%\\
    TPIa182I & 13.07\% & 17.91\% & 0.88 & 54.05\% & 82.49\%\\
    TPIa32I  & 13.56\% & 16.02\% & 0.88 & 55.76\% & 84.56\%\\
    TPIa33I  & 9.37\%  & 12.81\% & 0.91  & 67.13\% & 90.51\%\\
    \bottomrule
  \end{tabularx}
  \caption{Point-wise regression result of ZerNet under direction preserving setting over five validation cases.}
  \label{New_Prediction_table}
\end{table}

\begin{table}[h]
  \centering
  \begin{tabular}{l | r r r r}
    \toprule
    Angular resolution & $s=1$ & $s=4$ & $s=8$ & $s=16$ \\
    \midrule
    Training time & 2.08s & 7.55s & 14.58s & 28.26s\\
    Testing time & 1.55s & 4.55s & 8.26s & 15.52s\\
    \bottomrule
  \end{tabular}
  \caption{Computational time for ZerNet with different angular resolution.}
  \label{computing_time_table}
\end{table}

\section{Conclusion}
In this paper, we introduced a new concept of Zernike convolution as a way to generalize convolution to curved surfaces. We showed that Zernike convolution seamlessly generalizes convolution operations to an arbitrary surface in a mathematically rigorous and concise manner. In particular, we proved that manifold convolution can be formalized through decomposition of the local geometry defined on the tangent spaces and that convolution operations became simple dot products of Zernike polynomial coefficients. In addition, we showed that rotations of convolution kernels, which could be critical in manifold settings, could also be rigorously represented as simple $2 \times 2$ rotational transforms. Building upon this, we further demonstrated a promising vision of equipping our formulation as a theoretical foundation for direction-preserving convolution, which will bring more mathematical rigor to the generalization of geometric CNNs in fundamental. As a scalable algorithm developed upon our formulation, ZerNet also illustrated our competitiveness against other state-of-the-art methods on both classification and regression tasks.

For the future work, it would be worth exploring ways to further equip ZerNets with the other essential building blocks of CNNs such as spatial pooling/unpooling, transposed convolution, boundary padding etc. Particularly, encoder-decoder type networks on manifolds would be an interesting direction of research, as it can benefit potentially a large amount of computational geometry applications that requires parametrization (i.e. latent space embedding) of geometric shapes (e.g. \cite{Baek2012,shen2012detecting,Freifeld:ECCV:2012,Baek2013a,SMPL:2015,Baek2016b,pishchulin2017building}.


\begin{thebibliography}{10}

\bibitem{moosmann2009segmentation}
Frank Moosmann, Oliver Pink, and Christoph Stiller.
\newblock Segmentation of 3d lidar data in non-flat urban environments using a
  local convexity criterion.
\newblock In {\em Intelligent Vehicles Symposium, 2009 IEEE}, pages 215--220.
  IEEE, 2009.

\bibitem{douillard2011segmentation}
Bertrand Douillard, James Underwood, Noah Kuntz, Vsevolod Vlaskine, Alastair
  Quadros, Peter Morton, and Alon Frenkel.
\newblock On the segmentation of 3d lidar point clouds.
\newblock In {\em Robotics and Automation (ICRA), 2011 IEEE International
  Conference on}, pages 2798--2805. IEEE, 2011.

\bibitem{song2014sliding}
Shuran Song and Jianxiong Xiao.
\newblock Sliding shapes for 3d object detection in depth images.
\newblock In {\em European conference on computer vision}, pages 634--651.
  Springer, 2014.

\bibitem{Harik2017:ShapeTerra}
Ramy Harik, Yang Shi, and Stephen Baek.
\newblock Shape terra: mechanical feature recognition based on a persistent
  heat signature.
\newblock {\em Computer-Aided Design and Applications}, 14(2):206--218, 2017.

\bibitem{nurunnabi2015outlier}
Abdul Nurunnabi, Geoff West, and David Belton.
\newblock Outlier detection and robust normal-curvature estimation in mobile
  laser scanning 3d point cloud data.
\newblock {\em Pattern Recognition}, 48(4):1404--1419, 2015.

\bibitem{Zaharescu2009:meshHOG}
A.~Zaharescu, E.~Boyer, K.~Varanasi, and R.~Horaud.
\newblock Surface feature detection and description with applications to mesh
  matching.
\newblock In {\em 2009 IEEE Conference on Computer Vision and Pattern
  Recognition}, pages 373--380, June 2009.

\bibitem{Sun2017a}
Zhiyu Sun, Yusen He, Andrey Gritsenko, Amaury Lendasse, and Stephen Baek.
\newblock Deep spectral descriptors: Learning the point-wise correspondence
  metric via {S}iamese deep neural networks.
\newblock {\em {arXiv} Preprint: {arXiv:1710.06368}}, 10 2017.

\bibitem{rodola2014dense}
Emanuele Rodol{\`a}, Samuel Rota~Bulo, Thomas Windheuser, Matthias Vestner, and
  Daniel Cremers.
\newblock Dense non-rigid shape correspondence using random forests.
\newblock In {\em Proceedings of the IEEE Conference on Computer Vision and
  Pattern Recognition}, pages 4177--4184, 2014.

\bibitem{bronstein2017geometric}
Michael~M Bronstein, Joan Bruna, Yann LeCun, Arthur Szlam, and Pierre
  Vandergheynst.
\newblock Geometric deep learning: going beyond euclidean data.
\newblock {\em IEEE Signal Processing Magazine}, 34(4):18--42, 2017.

\bibitem{eisenberg1979proof}
Murray Eisenberg and Robert Guy.
\newblock A proof of the hairy ball theorem.
\newblock {\em The American Mathematical Monthly}, 86(7):571--574, 1979.

\bibitem{von1934beugungstheorie}
Zernike von F.
\newblock Beugungstheorie des schneidenver-fahrens und seiner verbesserten
  form, der phasenkontrastmethode.
\newblock {\em Physica}, 1(7-12):689--704, 1934.

\bibitem{bruna2013spectral}
Joan Bruna, Wojciech Zaremba, Arthur Szlam, and Yann LeCun.
\newblock Spectral networks and locally connected networks on graphs.
\newblock {\em arXiv preprint arXiv:1312.6203}, 2013.

\bibitem{henaff2015deep}
Mikael Henaff, Joan Bruna, and Yann LeCun.
\newblock Deep convolutional networks on graph-structured data.
\newblock {\em arXiv preprint arXiv:1506.05163}, 2015.

\bibitem{defferrard2016convolutional}
Micha{\"e}l Defferrard, Xavier Bresson, and Pierre Vandergheynst.
\newblock Convolutional neural networks on graphs with fast localized spectral
  filtering.
\newblock In {\em Advances in Neural Information Processing Systems}, pages
  3844--3852, 2016.

\bibitem{yi2017syncspeccnn}
Li~Yi, Hao Su, Xingwen Guo, and Leonidas~J Guibas.
\newblock {SyncSpecCNN}: Synchronized spectral {CNN} for {3D} shape
  segmentation.
\newblock In {\em IEEE Conference on Computer Vision and Pattern Recognition
  (CVPR)}, pages 2282--2290, 2017.

\bibitem{meyer2003discrete}
Mark Meyer, Mathieu Desbrun, Peter Schr{\"o}der, and Alan~H Barr.
\newblock Discrete differential-geometry operators for triangulated
  2-manifolds.
\newblock In {\em Visualization and Mathematics III}, pages 35--57. Springer,
  2003.

\bibitem{kipf2016semi}
Thomas~N Kipf and Max Welling.
\newblock Semi-supervised classification with graph convolutional networks.
\newblock {\em arXiv preprint arXiv:1609.02907}, 2016.

\bibitem{boscaini2015learning}
Davide Boscaini, Jonathan Masci, Simone Melzi, Michael~M Bronstein, Umberto
  Castellani, and Pierre Vandergheynst.
\newblock Learning class-specific descriptors for deformable shapes using
  localized spectral convolutional networks.
\newblock In {\em Computer Graphics Forum}, volume~34, pages 13--23. Wiley
  Online Library, 2015.

\bibitem{shuman2016vertex}
David~I Shuman, Benjamin Ricaud, and Pierre Vandergheynst.
\newblock Vertex-frequency analysis on graphs.
\newblock {\em Applied and Computational Harmonic Analysis}, 40(2):260--291,
  2016.

\bibitem{Levy:2010:SMP}
Bruno L{\'e}vy and Hao~(Richard) Zhang.
\newblock Spectral mesh processing.
\newblock In {\em ACM SIGGRAPH 2010 Courses}, 2010.

\bibitem{rong2008spectral}
Guodong Rong, Yan Cao, and Xiaohu Guo.
\newblock Spectral mesh deformation.
\newblock {\em The Visual Computer}, 24(7-9):787--796, 2008.

\bibitem{cohen2019gauge}
Taco~S Cohen, Maurice Weiler, Berkay Kicanaoglu, and Max Welling.
\newblock Gauge equivariant convolutional networks and the icosahedral {CNN}.
\newblock In {\em International Conference on Machine Learning (ICML)}, 2019.

\bibitem{duvenaud2015convolutional}
David~K Duvenaud, Dougal Maclaurin, Jorge Iparraguirre, Rafael Bombarell,
  Timothy Hirzel, Al{\'a}n Aspuru-Guzik, and Ryan~P Adams.
\newblock Convolutional networks on graphs for learning molecular fingerprints.
\newblock In {\em Advances in neural information processing systems}, pages
  2224--2232, 2015.

\bibitem{niepert2016learning}
Mathias Niepert, Mohamed Ahmed, and Konstantin Kutzkov.
\newblock Learning convolutional neural networks for graphs.
\newblock In {\em International conference on machine learning}, pages
  2014--2023, 2016.

\bibitem{hechtlinger2017generalization}
Yotam Hechtlinger, Purvasha Chakravarti, and Jining Qin.
\newblock A generalization of convolutional neural networks to graph-structured
  data.
\newblock {\em arXiv preprint arXiv:1704.08165}, 2017.

\bibitem{masci2015geodesic}
Jonathan Masci, Davide Boscaini, Michael Bronstein, and Pierre Vandergheynst.
\newblock Geodesic convolutional neural networks on riemannian manifolds.
\newblock In {\em Proceedings of the IEEE international conference on computer
  vision workshops}, pages 37--45, 2015.

\bibitem{boscaini2016learning}
Davide Boscaini, Jonathan Masci, Emanuele Rodol{\`a}, and Michael Bronstein.
\newblock Learning shape correspondence with anisotropic convolutional neural
  networks.
\newblock In {\em Advances in Neural Information Processing Systems}, pages
  3189--3197, 2016.

\bibitem{monti2017geometric}
Federico Monti, Davide Boscaini, Jonathan Masci, Emanuele Rodola, Jan Svoboda,
  and Michael~M Bronstein.
\newblock Geometric deep learning on graphs and manifolds using mixture model
  cnns.
\newblock In {\em Proc. CVPR}, volume~1, page~3, 2017.

\bibitem{hanocka2019meshcnn}
Rana Hanocka, Amir Hertz, Noa Fish, Raja Giryes, Shachar Fleishman, and Daniel
  Cohen-Or.
\newblock {MeshCNN}: A network with an edge.
\newblock In {\em ACM SIGGRAPH}, 2019.

\bibitem{masci2016geometric}
Jonathan Masci, Emanuele Rodol{\`a}, Davide Boscaini, Michael~M Bronstein, and
  Hao Li.
\newblock Geometric deep learning.
\newblock In {\em SIGGRAPH ASIA 2016 Courses}, page~1. ACM, 2016.

\bibitem{verma2018feastnet}
Nitika Verma, Edmond Boyer, and Jakob Verbeek.
\newblock Feastnet: Feature-steered graph convolutions for 3d shape analysis.
\newblock In {\em CVPR 2018-IEEE Conference on Computer Vision \& Pattern
  Recognition}, 2018.

\bibitem{poulenard2018multi}
Adrien Poulenard and Maks Ovsjanikov.
\newblock Multi-directional geodesic neural networks via equivariant
  convolution.
\newblock {\em ACM Transactions on Graphics (TOG)}, 37(6):Article No. 236,
  2018.

\bibitem{schonsheck2018parallel}
Stefan~C Schonsheck, Bin Dong, and Rongjie Lai.
\newblock Parallel transport convolution: A new tool for convolutional neural
  networks on manifolds.
\newblock {\em arXiv preprint arXiv:1805.07857}, 2018.

\bibitem{schmidt2006interactive}
Ryan Schmidt, Cindy Grimm, and Brian Wyvill.
\newblock Interactive decal compositing with discrete exponential maps.
\newblock {\em ACM Transactions on Graphics (TOG)}, 25(3):605--613, 2006.

\bibitem{melvaer2012geodesic}
Eivind~Lyche Melv{\ae}r and Martin Reimers.
\newblock Geodesic polar coordinates on polygonal meshes.
\newblock {\em Computer Graphics Forum}, 31(8):2423--2435, 2012.

\bibitem{Crane:2017:HMD}
Keenan Crane, Clarisse Weischedel, and Max Wardetzky.
\newblock The heat method for distance computation.
\newblock {\em Communications of the ACM}, 60(11):90--99, 2017.

\bibitem{Bogo:CVPR:2014}
Federica Bogo, Javier Romero, Matthew Loper, and Michael~J. Black.
\newblock {FAUST}: Dataset and evaluation for {3D} mesh registration.
\newblock In {\em Proceedings IEEE Conf. on Computer Vision and Pattern
  Recognition (CVPR)}, Piscataway, NJ, USA, June 2014. IEEE.

\bibitem{boscaini2016anisotropic}
Davide Boscaini, Jonathan Masci, Emanuele Rodol{\`a}, Michael~M Bronstein, and
  Daniel Cremers.
\newblock Anisotropic diffusion descriptors.
\newblock In {\em Computer Graphics Forum}, volume~35, pages 431--441. Wiley
  Online Library, 2016.

\bibitem{DBLP:journals/corr/KingmaB14}
Diederik~P. Kingma and Jimmy Ba.
\newblock Adam: {A} method for stochastic optimization.
\newblock {\em CoRR}, abs/1412.6980, 2014.

\bibitem{Ovsjanikov:2012}
Maks Ovsjanikov, Mirela Ben-Chen, Justin Solomon, Adrian Butscher, and Leonidas
  Guibas.
\newblock Functional maps: A flexible representation of maps between shapes.
\newblock {\em ACM Trans. Graph.}, 31(4):30:1--30:11, July 2012.

\bibitem{vestner2017product}
Matthias Vestner, Roee Litman, Emanuele Rodol{\`a}, Alexander~M Bronstein, and
  Daniel Cremers.
\newblock Product manifold filter: Non-rigid shape correspondence via kernel
  density estimation in the product space.
\newblock In {\em CVPR}, pages 6681--6690, 2017.

\bibitem{kim2011blended}
Vladimir~G Kim, Yaron Lipman, and Thomas Funkhouser.
\newblock Blended intrinsic maps.
\newblock In {\em ACM Transactions on Graphics (TOG)}, volume~30, page~79. ACM,
  2011.

\bibitem{maron2017convolutional}
Haggai Maron, Meirav Galun, Noam Aigerman, Miri Trope, Nadav Dym, Ersin Yumer,
  Vladimir~G Kim, and Yaron Lipman.
\newblock Convolutional neural networks on surfaces via seamless toric covers.
\newblock {\em ACM Trans. Graph.}, 36(4):71--1, 2017.

\bibitem{qi2017pointnet++}
Charles~Ruizhongtai Qi, Li~Yi, Hao Su, and Leonidas~J Guibas.
\newblock Pointnet++: Deep hierarchical feature learning on point sets in a
  metric space.
\newblock In {\em Advances in neural information processing systems}, pages
  5099--5108, 2017.

\bibitem{wang2018dynamic}
Yue Wang, Yongbin Sun, Ziwei Liu, Sanjay~E Sarma, Michael~M Bronstein, and
  Justin~M Solomon.
\newblock Dynamic graph cnn for learning on point clouds.
\newblock {\em arXiv preprint arXiv:1801.07829}, 2018.

\bibitem{Adobe2016}
Adobe.
\newblock Adobe fuse 3d characters.
\newblock 2016.

\bibitem{lu2007inverse}
Jia Lu, Xianlian Zhou, and Madhavan~L Raghavan.
\newblock Inverse elastostatic stress analysis in pre-deformed biological
  structures: demonstration using abdominal aortic aneurysms.
\newblock {\em Journal of biomechanics}, 40(3):693--696, 2007.

\bibitem{lu2008inverse}
Jia Lu, Xianlian Zhou, and Madhavan~L Raghavan.
\newblock Inverse method of stress analysis for cerebral aneurysms.
\newblock {\em Biomechanics and modeling in mechanobiology}, 7(6):477--486,
  2008.

\bibitem{lu2016solving}
Jia Lu and Yuanming Luo.
\newblock Solving membrane stress on deformed configuration using inverse
  elastostatic and forward penalty methods.
\newblock {\em Computer Methods in Applied Mechanics and Engineering},
  308:134--150, 2016.

\bibitem{luo2018machine}
Yuanming Luo, Zhiwei Fan, Stephen Baek, and Jia Lu.
\newblock Machine learning--aided exploration of relationship between strength
  and elastic properties in ascending thoracic aneurysm.
\newblock {\em International journal for numerical methods in biomedical
  engineering}, 34(6):e2977, 2018.

\bibitem{Baek2012}
Seung-Yeob Baek and Kunwoo Lee.
\newblock Parametric human body shape modeling framework for human-centered
  product design.
\newblock {\em Computer-Aided Design}, 44(1):56--67, 1 2012.

\bibitem{shen2012detecting}
Kai-kai Shen, Jurgen Fripp, Fabrice M{\'e}riaudeau, Ga{\"e}l Ch{\'e}telat,
  Olivier Salvado, Pierrick Bourgeat, Alzheimer's Disease~Neuroimaging
  Initiative, et~al.
\newblock Detecting global and local hippocampal shape changes in alzheimer's
  disease using statistical shape models.
\newblock {\em Neuroimage}, 59(3):2155--2166, 2012.

\bibitem{Freifeld:ECCV:2012}
Oren Freifeld and Michael~J. Black.
\newblock Lie bodies: A manifold representation of {3D} human shape.
\newblock In {\em European Conf. on Computer Vision (ECCV)}, Part I, LNCS 7572,
  pages 1--14. Springer-Verlag, October 2012.

\bibitem{Baek2013a}
Seung-Yeob Baek, Joon~Ho Wang, Insub Song, Kunwoo Lee, Jehee Lee, and Seungbum
  Koo.
\newblock Automated bone landmarks prediction on the femur using anatomical
  deformation technique.
\newblock {\em Computer-Aided Design}, 45(2):505--510, 2 2013.

\bibitem{SMPL:2015}
Matthew Loper, Naureen Mahmood, Javier Romero, Gerard Pons-Moll, and Michael~J.
  Black.
\newblock {SMPL}: A skinned multi-person linear model.
\newblock {\em ACM Trans. Graphics (Proc. SIGGRAPH Asia)}, 34(6):248:1--248:16,
  October 2015.

\bibitem{Baek2016b}
Seung-Yeob Baek and Kunwoo Lee.
\newblock Statistical foot-shape analysis for mass-customisation of footwear.
\newblock {\em International Journal of Computer Aided Engineering and
  Technology}, 8(1/2):80--98, 1 2016.

\bibitem{pishchulin2017building}
Leonid Pishchulin, Stefanie Wuhrer, Thomas Helten, Christian Theobalt, and
  Bernt Schiele.
\newblock Building statistical shape spaces for 3d human modeling.
\newblock {\em Pattern Recognition}, 67:276--286, 2017.

\end{thebibliography}

\end{document}